\documentclass[runningheads]{llncs}
\usepackage{graphicx}
\usepackage{amsmath,amssymb} 
\usepackage{color}
\usepackage{url}
\usepackage[width=122mm,left=12mm,paperwidth=146mm,height=193mm,top=12mm,paperheight=217mm]{geometry}
\begin{document}

\newcommand{\iancomment}[1]{{\color{red} IAN: #1}}
\newcommand{\ravicomment}[1]{{\color{blue} RAVI: #1}}
\newcommand{\anderscomment}[1]{{\color{green} ANDERS: #1}}

\pagestyle{headings}
\mainmatter
\title{Unsupervised CNN for Single View Depth Estimation: Geometry to the Rescue}
\titlerunning{Unsupervised CNN: Geometry to the Rescue}
\author{Ravi Garg, Vijay Kumar B G, Gustavo Carneiro, Ian Reid}
\institute{The University of Adelaide, SA 5005, Australia\\
	\email{ \{ravi.garg,vijay.kumar,gustavo.carneiro,ian.reid\}@adelaide.edu.au}
}
\authorrunning{Ravi Garg, Vijay Kumar B G, Gustavo Carneiro, Ian Reid }
\maketitle

\begin{abstract}
A significant weakness of most current deep Convolutional Neural Networks is the need to train them using vast amounts of manually labelled data.   
In this work we propose a unsupervised framework to learn a deep convolutional neural network for single view depth prediction, without requiring a pre-training stage 
or annotated ground-truth depths.  We achieve this by training the network in a manner analogous to an autoencoder.  At training time we consider a pair of images, source and target, with small, known camera motion between the two such as a stereo pair.  We train the convolutional encoder for the task of predicting the depth map for the source image.  To do so, we explicitly generate an inverse warp of the target image using the predicted depth and known inter-view displacement, to reconstruct the source image; the photometric error in the reconstruction is the reconstruction loss for the encoder. The acquisition of this training data is considerably simpler than for equivalent systems, requiring no manual annotation, nor calibration of depth sensor to camera.  
%
%
%
We show that our network trained on less than half of the KITTI dataset gives comparable
performance to that of the state-of-the-art supervised methods for single view depth estimation. \footnote{Find the model and other imformation on the project github page: \url{https://github.com/Ravi-Garg/Unsupervised_Depth_Estimation}} 
\end{abstract}

\begin{figure*}[!t]
\centering\includegraphics[width=1.0\textwidth]{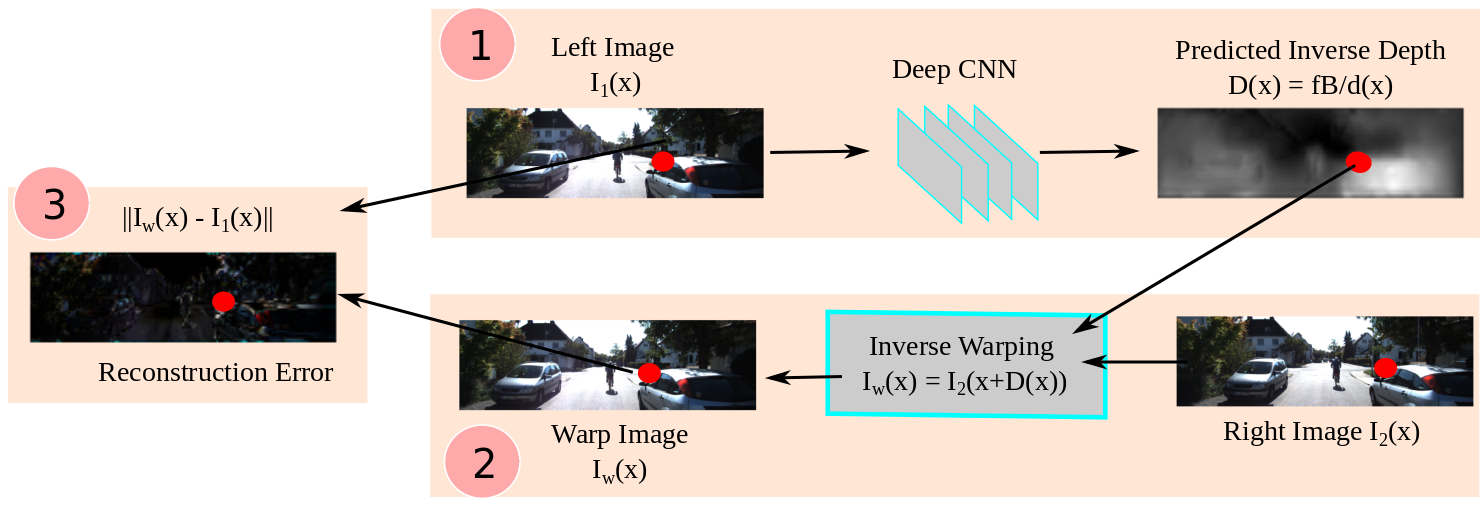} 
\caption{We propose a stereopsis based auto-encoder setup: the encoder (Part 1)
is a traditional convolutional neural network with stacked convolutions and pooling layers (See Figure \ref{fig:FCN}) and maps the left image ($I_1$) of the rectified stereo pair into its depth map.
Our decoder (Part 2) explicitly forces the encoder output to be disparities (scaled inverse depth) by synthesizing a backward warp image ($I_w$) by moving pixels
from right image $I_2$ along the scan-line. We use the reconstructed output $I_w$ to be matched with the encoder input (Part 3) via a simple loss. For end-to-end training, we minimize the reconstruction 
loss with a simple smoothness prior on disparities which deals with the aperture problem, while at test time our CNN performs single-view disparity (inverse depth) prediction, up to the scene 
scale given in form of $fB$ at the time of training.\label{fig:idea}}
\end{figure*}

\section{Introduction}\label{sec:intro}

The availability of very large human annotated datasets like Imagenet \cite{imagenet_cvpr2009} has led to a surge of deep learning approaches successfully addressing various vision problems.
Trained initially on tasks such as image classification, and fine-tuned to fit other tasks, supervised CNNs are now state-of-the-art 
for object detection \cite{girshickCVPR2014}, per-pixel image classification \cite{nohfcnICCV2015}, depth and normal prediction from single image \cite{chunhuaCVPR2015}, human pose estimation \cite{FanCVPR2015} and many other applications.
A significant and abiding weakness, however, is the need to accrue labeled data for the supervised learning. Providing per-pixel segmentation
masks on large datasets like CoCo \cite{cocoeccv2014}, or classification labels for Imagenet requires significant human effort and is prone to error.  
Supervised training for single view depth estimation for outdoor scenes requires expensive hardware and careful acquisition\cite{make3dPAMI2009,DeptheigenNIPS2014,DepthLiuPAMI2016,Ladicky_2014_CVPR}.
For example, despite using state-of-the-art 3D sensors, multiple calibrated cameras and inertial sensors, a dataset like KITTI \cite{GeigerIJRR2013} 
provides sparse depthmaps with less than 5\% density on the captured image resolutions and with only a limited reliable depth range. 
A significant challenge now is to develop unsupervised training regimes that can train networks that perform either as well as,
or better than those trained used using these supervised methods.  This will be a major step towards realizing in-situ learning,
in which we can retrain or tune a network for specific circumstances, and towards life-long learning, in which continuous acquisition of data leads to improved performance over time.

In this paper we are particularly concerned with the task of single-view depth estimation, in which the goal is to learn a non linear prediction function which maps an image to its depth map.  CNNs have achieved the state-of-the-art performance on this task due to their ability to capture the complex and implicit relationships between scene depth and the corresponding image textures, scene semantics, and local and global context in the image. State-of-the-art supervised learning methods for this task train a CNN to minimize a loss based on either the scale invariant RMS \cite{DeptheigenNIPS2014}, or the log RMS \cite{DepthLiuPAMI2016} of the depth predictions from ground-truth. These networks have been trained using datasets that provide both RGB images and corresponding depthmaps such as NYUv2 and KITTI. 

However as noted in \cite{DepthLiuPAMI2016}, the networks learned by these systems do not generalize well outside their immediate domain of application.  For example, \cite{DepthLiuPAMI2016} trained two separate networks, one for indoors (using NYUv2) and one for street scenes (using KITTI), because the weights learned in one do not work well in the other.  To transfer the idea of single-view depth estimation into yet another domain would require indulging in the expensive task of acquiring a new RGB-D dataset with well aligned image and depth values, and re-train the network.  An alternative to this would be to generate a large synthetic or semi-synthetic dataset using graphical rendering, an approach that has met with some success in \cite{HandaPBSC15a}.  However it is difficult to capture the full variability of real-world images in such datasets. 

Another possible approach would be to capture a large dataset of stereo images, and use standard geometric methods to compute the disparity map for each pair, yielding a large set of image-plus-disparity-map pairs.  We could then train a network to predict a disparity map from a single view.  However such system will likely learn the
systematic errors in estimated depths, ``baking in'' the failure modes of the stereo algorithm. Factors such as sensor flare, motion blur, lighting changes, shadows, etc are
present in real images and rarely dealt with adequately by standard stereo algorithms.

We adopt a different approach that moves towards a system capable of in-situ training or even lifelong learning, using real un-annotated imagery.  We take inspiration from the idea of autoencoders, and leverage well-understood ideas in visual geometry. The result is a convolutional neural network for single-view depth estimation, the first of its kind that can be trained end-to-end from scratch, in a fully unsupervised fashion, simply using data captured using a stereo rig.  

\section{Approach}

In this section we give more detail of our approach.  Figure \ref{fig:idea} explains our idea graphically. 
To train our network, we make use of pairs of images with a known camera motion between the two, such as stereo pairs.  Such data are considerably more easily acquired than calibrated depthmaps and aligned images.  In our case we use large numbers of stereo pairs, but the method applies equally to data acquired from a moving SLAM system in an otherwise static scene.  

We learn a CNN to model the complex non-linear transformation which converts the image to a depth-map.  The loss we use for learning this CNN is the photometric difference between the input -- or source -- image, and the inverse-warped target image (the other image in the stereo pair).  This loss is both differentiable (to facilitate back-propagation) and is highly correlated with the prediction error - i.e. can be used to accurately rank two different depth-maps without using ground-truth labels.

This approach can be interpreted in the context of convolutional autoencoders. The task of the standard autoencoder is to encode the input with a series of non-linear operations to a compressed code that captures sufficient core information so that a decoder can reconstruct the input with minimal reconstruction error.  In our case we replace the decoder with a standard geometric image warp, based on the predicted depth map and the relative camera positions. 
This has two advantages: first, the decoder in our case does not need to be learned, since it is already a well-understood geometric operation; second, our reconstruction loss naturally encourages the code to be the correct depth image. 

\subsection{Autoencoder loss}
Every training instance $i \in \{1 \cdots N\}$ in our setup is a rectified stereo pair $\{ I_1^i ,I_2^i \}$ captured by a single pre-calibrated stereo rig with two 
cameras having focal length $f$ each which are separated horizontally by a distance $B$.\footnote{All training images are assumed to be taken with a fixed rectified
stereo setup as is the case in KITTI for simplicity but our method is generalizable to work with instances taken by different calibrated stereos.}
Assuming that the predicted depth of a pixel $x$ for the left image of the rig via CNN is $d^i(x)$, the motion of the pixel along the scan-line $D^i(x)$ is then 
$fB/d^i(x)$. Thus, using the right image $I_2^i$, a warp $I_w^i$ can be synthesized as $I_2^i(x+fB/d^i(x))$. 

With this explicit parameterization of the warp, we propose to minimize standard color constancy (photometric) error between the reconstructed image $I_w^i$ and the left image $I_1^i$:
\begin{equation}
E^i_{recons} = \int_\Omega \|I_w^i(x) - I_1^i(x)\|^2 dx = \int_\Omega \| I_2^i(x + \underbrace{D^i(x)}_{fB/d^i(x)}) -I_1^i(x) \|^2 dx \label{eq:Edata}
\end{equation}
It is well known that this photometric loss function is non-informative in homogeneous regions of the scene.
Thus multiple disparities can generate
equally good warps $I_w$'s and a prior on the disparities is needed to get a unique depthmap. 
We use very simple $L2$ regularization on the disparity discontinuities as our prior to deal with the aperture problem:
\begin{equation}
E^i_{smooth} = \| \nabla D^i(x) \|^2
\end{equation}
This regularizer is known to over-smooth the estimated motion, however a vast literature of more sophisticated edge preserving regularizers
with robust penalty functions  like \cite{brox2004high,zach2007duality} for which gradients can be computed are at our 
disposal and can be easily used with our setup to get sharper depthmaps. 
As the main purpose of our work is to prove that end-to-end training of the proposed autoencoder is feasible and helpful for depth prediction,
we choose to minimize the simplest suitable loss summed over all training instances:
\begin{equation}
E = \sum\limits_{i=1}^N E^i_{recons}  + \gamma E^i_{smooth} \label{eq:LossHS}
\end{equation}
where $\gamma$ is the strength of the regularization forcing the estimated depthmaps to be smooth.

Our loss function as described in \eqref{eq:LossHS} is similar to the standard Horn and Schunck optic flow cost \cite{horn1981determining}  for every frame.
However, the major difference is that our disparity maps  $D^i$'s are parametrized to be a non-linear function of the input image and unknown weights of the CNN
which are shared for estimating the motion between every stereo pair. This parameter sharing enforces consistency in the estimated depths over 1000's of correlated training images
of a large dataset like KITTI.
Our autoencoder's reconstruction loss can be seen as a generalization of the multi-frame optic flow methods like \cite{garg2013dense,garg2010dense}. The difference is, 
instead of modeling the correlations in the estimated motions for a shorter video sequence with a predefined linear subspace \cite{garg2010dense},
our autoencoder learns (and models) valid flows which are consistent throughout the dataset non-linearly.

\begin{figure}[!t]
\centering\includegraphics[width=.75\textwidth]{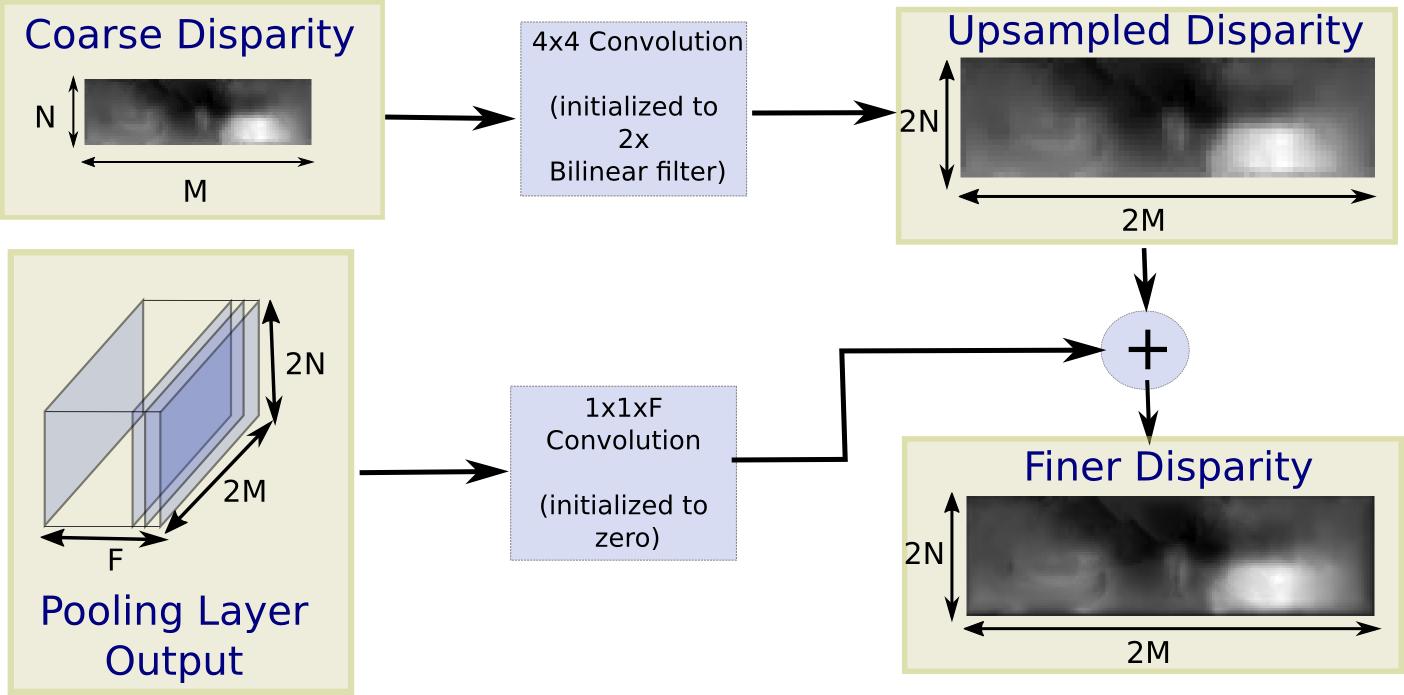} 
\caption{Coarse-to-fine stereo with CNN with results on a  sample validation instance: 
We adapt the convolution based upsampling architecture proposed in \cite{FullyconvolutionalCVPR2015} to mimic
the coarse-to-fine stereo estimations. Our upsampling filter is initialized with simple
bilinear interpolation kernel and we initialize the corresponding pooling layer contribution by setting both bias
and 1x1 convolution filter to be zero. The figure shows how features coming from previous layers of the CNN (L3)
combined with finer resolution loss function generate better depthmaps at $44\times 172$ from our bilinear
upsampled initial estimate of coarser prediction at $22\times 76$.\label{fig:FCN}}
\end{figure}

\section{Coarse-to-fine training with skip architecture} \label{sec:coarse_to_fine}
To compute the gradient for standard back-propagation on our cost \eqref{eq:Edata},
we need to linearize the warp image at the current estimate of the disparities using Taylor expansion:
\begin{equation}
I_2(x + D^{n}(x)) = I_2(x+D^{n-1}(x)) + (D^{n}(x) - D^{n-1}(x)) I_{2h}(x+D^{n-1}(x))
\end{equation}
where $I_{2h}$ represents the horizontal gradient of the warp image computed at the current disparity $D^{n-1}$ at iteration $n$.\footnote{We have dropped the training instance index $i$ for simplicity.}  
This linearization is valid only for small values of $D^{n}(x) - D^{n-1}(x)$ limiting the magnitude of estimated disparities in the image.
To estimate larger motions (smaller depths) accurately, a coarse-to-fine strategy with iterative warping is well established in the 
stereo and optic flow literature which facilitates gradient descent-based continuous optimization. 
We refer the readers to \cite{steinbrucker2009large} for more detailed discussion of the requirements of this linearization, its limitations and existing alternatives.

However, our disparities are a non-linear function of the CNN parameters and the input image. 
To move from coarse-to-fine level, we not only need a good disparity initialization at the 
finer resolutions to linearize the warps but also the corresponding CNN parameters which predict
these initial disparities for each training instance.  Fortunately, the recent fully-convolutional
architecture with upsampling, proposed in \cite{FullyconvolutionalCVPR2015}, is a suitable choice to enable coarse-to-fine
warping for our system. As depicted in Figure \ref{fig:FCN}, given a network which predicts an $M\times N$
disparities, we can use a simple bilinear upsampling filter to initialize upscaled disparities 
(to get $2M\times2N$ depthmaps) keeping the other network parameters fixed. 
It has been shown that the
finer details of the images are captured in the previous layers of CNN, and fusing back such information 
is helpful for refining a coarse CNN prediction. We use $1\times 1$ convolution with the filter and bias both initialized to zero and the convolved output is then
fused with the upscaled depths with an element-wise sum layer for refinement.

\section{Network Architecture}

\begin{figure}[!t]
\centering
\includegraphics[width=1.0\textwidth]{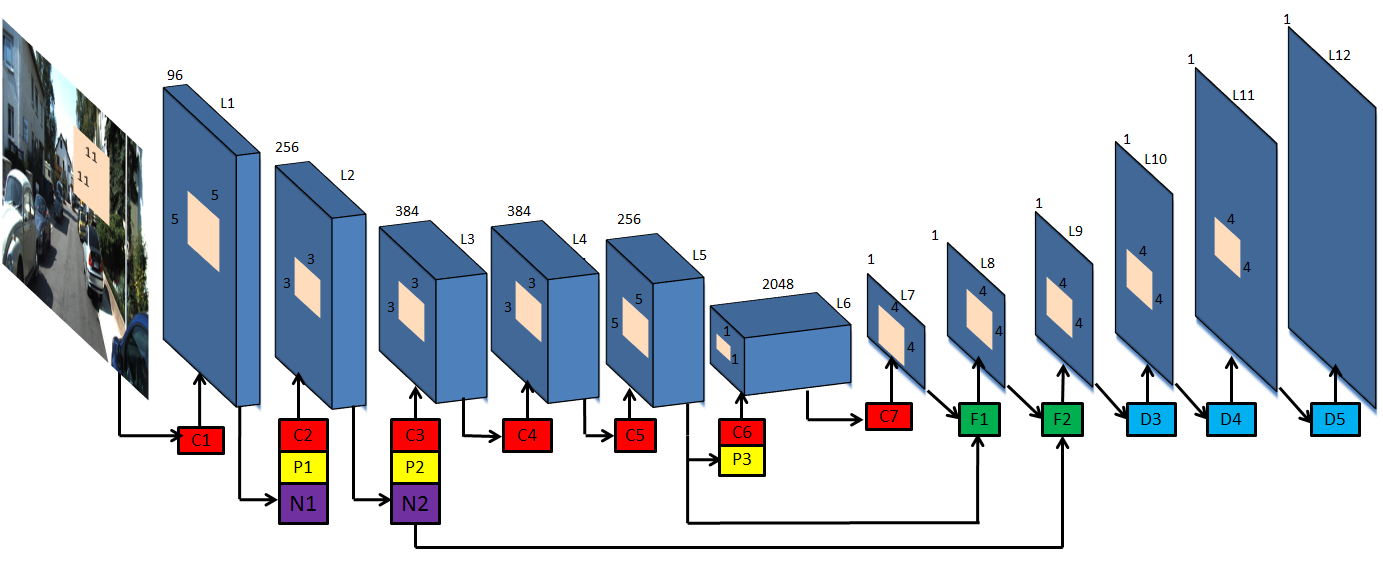}
\caption{Network architecture: The blocks C (red), P (yellow), L (dark blue), F (green), D (blue) correspond to convolution, pooling, local response normalization, FCN and upsampling layers respectively. 
The FCN blocks F1 and F2 upsample the predictions from layers (L7, L8) and combine it with the input of the pooling layers P3 and P2 respectively.\label{fig:net_arch}}
\end{figure}

\noindent The network architecture for our deep convolutional encoder is shown in Figure \ref{fig:net_arch} which is similar to the Alexnet architecture \cite{AleximagenetNIPS2012} up to the C5 layer.
We replace the fully connected layer of Alexnet by a fully convolutional layer with 2048 convolution filters 
of size $5 \times 5$ each.\footnote{A $5 \times 18$ convolution can be used instead to increase network capacity and replicate the effect of a fully connected layer of \cite{AleximagenetNIPS2012}.}
This reduces the number of parameters in the network and allows for the network to accept variable size inputs at test time. 
More importantly, it preserves the spatial information present in the image and allows us to upsample the predictions in a stage-wise manner in the layers that follow the L7 output of the figure, which is a requirement for our stereopsis based autoencoder.
 Inspired by the observations from \cite{FullyconvolutionalCVPR2015}, that the finer details in the images are lost in the last few layers of the deep convolutional network we 
employ the ``skip architecture'' that combines the coarser depth prediction with the local image information to get finer predictions. The effect of this is illustrated using an 
example from the validation set in Figure \ref{fig:FCN}. 
The layers following the L9 output ($22\times 76$ depthmap) in our network are simple $4\times 4$ convolutions 
each converting a coarser low resolution depth map to a higher resolution output.

\section{Experiments}
We evaluate our method on the publicly available KITTI dataset \cite{GeigerIJRR2013} that comprises several outdoor scenes captured using a stereo camera mounted on a moving vehicle.
We employ the same train/test split used in \cite{DeptheigenNIPS2014}: from the 56 scenes belonging to the categories ``city'', ``residential'' and ``road'', we choose 28 for training and 
the remaining 28 for testing. We downsample the left images by a factor of 2 to bring them to 188 x 620, and at this resolution they are used as input to the network. 
Each corresponding right image in a stereo pair is used at the resolution of the predicted depthmap at each stage of our coarse-to-fine training to 
generate the warp and match it with a resized left image.

The training set consists of 23488 stereo pairs out of which we use 22600 for training and the remaining for validation. Neither the right to left stereo 
nor any data augmentation are used for the coarse-to-fine training in multiple stages. 
For testing, we use the 697 images provided by \cite{DeptheigenNIPS2014}. We do not use any ground-truth depths for training the network.
To evaluate all the results produced by our network we use simple upscaling of the low resolution disparity predictions to the resolution at which the stereo images were captured.
Using the stereo baseline of $0.54$ metres, we convert the upsampled disparities to generate depthmaps at KITTI resolution using  $d = fB/D$.

For fair comparison with state-of-the-art single view depth prediction, we evaluate our results on the same cropped region of interest as \cite{DeptheigenNIPS2014}. 
Since the supervised methods are trained using the ground-truth depth that ranges between $1$ and $50$ meters whereas we can predict larger depths, we 
clamp the predicted depth values for our method between $1$ and $50$ for evaluation. i.e. setting the depths bigger than 50 metres to 50.
\noindent We evaluate our method using the error measures reported in \cite{DeptheigenNIPS2014,DepthLiuPAMI2016}:%


\begin{tabular}{l l l}
{{RMS}:\scalebox{0.8}{ $\sqrt{\frac{1}{T}\sum_{i \in T}\lVert d_i - d_i^{gt} \rVert^2}$} }  
& \ \ \ \ \ \ \ 
& {{log RMS}: \scalebox{0.8}{$\sqrt{\frac{1}{T}\sum_{i \in T}\lVert log(d_i) - log(d_i^{gt}) \rVert^2}$}}\\
{{abs. relative: $\frac{1}{T}\sum_{i \in T}\frac{\lvert d_i - d_i^{gt} \rvert}{d_i^{gt}}$}} 
& \ \ \ \ \ \ \ 
&{{sq. relative: $\frac{1}{T}\sum_{i \in T}\frac{\lVert d_i - d_i^{gt} \rVert^2}{d_i^{gt}}$}}\\
\multicolumn{3}{l}{{Accuracies:  $\%$ of $ d_i $ s.t. $max(\frac{d_i}{d_i^{gt}},\frac{d_i^{gt}}{d_i}) = \delta < thr$}}
\end{tabular}

\subsection{Implementation Details}
We train our network using the CNN toolbox MatConvnet \cite{vedaldi15matconvnet}. We use SGD for optimization with momentum $0.9$ and weight decay of $0.0005$.
Our network weights are initialized randomly for the first 5 layers of the Alexnet and we append the $5\times5$ 
fully convolutional layer initialized with zero weights to get zero disparity estimates. We subtract every pixel's color by 128 and divide it by 255 to have both left and right images $\in [-0.5,0.5]$.
The smoothness prior strength $\gamma$ was set to $0.01$.

Due to the linearization of the loss function as explained in Section \ref{sec:coarse_to_fine}, we learn the network proposed in Figure \ref{fig:net_arch} in multiple stages, starting from the coarsest level (L7 in Figure \ref{fig:net_arch}),
and iteratively adding upsampling layers one at a time.  The learning rate for the network which predicts depths at the coarsest resolution is initialized to $0.01$ and gradually decreased after each epoch using the factor $1/(1+\alpha *n)^{(n-1)}$
where $n$ is the index of current epoch and $\alpha = 0.0005$. The smoothness prior strength $\gamma$ was set to $0.01$.
We train this coarse depth prediction network (L1-L7) for 100 epochs.

\begin{table}[!t]
\centering
\caption{Performance of the proposed framework at various stages of training. } \label{tab:coarse_to_fine}
\begin{scriptsize}
 \begin{tabular}{|c |c|c|c|c|c|c|c|c|c|c|c|c|c|}
 \hline
		    & &  &$log$  & Absolute &Square  &&Accuracies& \\ 
 Methods &Resolution&RMS&RMS& relative& relative&$\delta < 1.25$ &$\delta < 1.25^2$ &$ \delta <1.25^3$\\ \hline
 Ours  L9 	&$22\times 76$	&5.740 &0.310  &0.205 &1.353 		&0.660 &0.872 &0.948
 
 \\ \hline
 Ours L10 + skip\footnotemark  	&$46\times 154$	&5.850 &0.338  &0.246 &1.673		&0.607 &0.842 &0.937
 \\ \hline 
 
 Ours  L10 	&$44\times 152$	&5.434 &0.292  &0.189 &1.214		&0.705 &0.889 &0.955

 \\ \hline
 
 Ours  L11	&$88\times 304$	&{5.326} &{0.285}  &{0.179} &{1.177}		&{0.721} &{0.892} &{0.958}

  \\ \hline 
 Ours  L12	&$176\times 608$	&{ 5.285} &{0.282}  &{0.177} &{ 1.169}		&{0.727} &{0.896} &{0.958}
  \\ \hline \hline
 Ours  L12, Aug. 8x	&	&{ 5.104} &{0.273}  &{0.169} &{ 1.08}		&{0.740} &{0.904} &{0.962}
 \\ \hline
 \end{tabular}
 \end{scriptsize}
\end{table}
\footnotetext{Layer 10 result while using 3rd skip-connection.}

\begin{figure}[!t]
\includegraphics[width=1\textwidth]{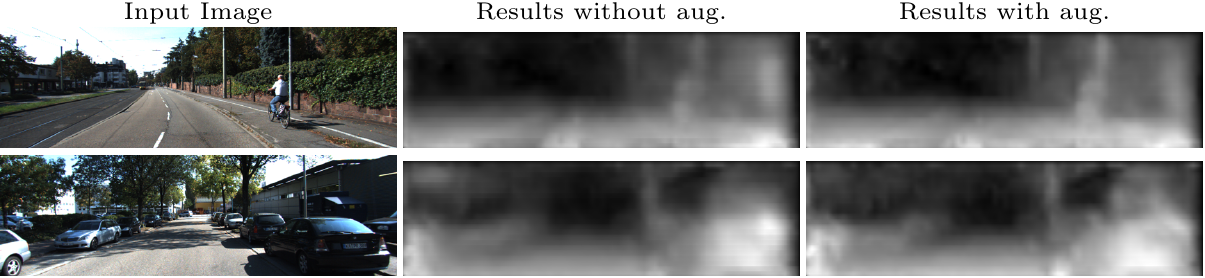} 
\caption{Data augmentation improves the predicted disparities for smaller objects. Look at the biker in the first and the bottom right car in the second example.\label{fig:visualization_aug}}
\end{figure}

\subsection{Effect of upsampling}
Having the coarser depth estimates for the training-set, we iteratively add upsampling layers which increases the resolution of the predictions by a factor of $\approx 2$.
\footnote{Alexnet uses uneven padding for some convolutions leading to change in the aspect ratio and the image size.}
Since the number of pixels in the images are increased by a factor of $4$, the cost approximately increases by the same factor when moving from coarser to finer level training. 
Hence we decrease the initial learning rate by a factor of $4$ for training the finer networks.  
Starting from the coarsest predictions (L7) we progressively add upsampling layers L8 to L12 to get depths at resolutions
$10 \times 37$, $22 \times 76$, $44 \times 152$, $88 \times 304$ and $176 \times 608$ respectively. We train each of the finer networks for 100 
epochs with the decaying learning rate as described in previous section.
While adding the upsampling layers, we crop and pad the layers such that the resolution of predictions in L8 and L9
matches the resolution of the input to the pooling layers P3 and P2 respectively. 
For the upsampling layers without skip-connection padding of 1 pixel is used.

Table \ref{tab:coarse_to_fine} analyses the disparity estimation accuracy for our network on the KITTI test-set at various stages of the training.
Row 1 and 2 of our table correspond to our L9 and L10 output with 2 and 3 FCN blocks respectively. 
Consistent with \cite{FullyconvolutionalCVPR2015} we also observe that after 2 upsampling layers, 
the skipped architecture starts to give diminishing returns. As evident from the third row in Table \ref{tab:coarse_to_fine} layer L10 without skip-connection outperforms the counterpart.
We believe that this is due to the fact that the features learned in the first few layers of the CNN are more relevant to ordinary photometric images than to the depth images. Thus, a simple weighted sum of
these features with that of the depth map does not work well. However, higher resolution images still have richer information for image correspondences which can be back-propagated via
our loss function for better predictions. The gradual improvement in disparity estimations using high resolution images is evident in Table \ref{tab:coarse_to_fine}.

\subsection{Fine tuning with augmentation}

Once we have our base network trained in the stage-wise manner described above, we further fine-tune this network (without coarse-to-fine training) 
for another 100 epochs with following augmentations:
\begin{itemize}
 \item Color ($2\times$): Color channels are multiplied by a factor $c \in [0.9, 1.1]$ randomly.
 \item Scale ($2\times$): We scale the input image by a factor of $s \in [1,1.6]$ and randomly crop the images to match the network input size.
 \item Left-Right flips  ($2\times$): We flip left and right images horizontally and swap them to get new training pair with positive disparities to keep consistency.
\end{itemize}
Consistent with other CNNs, fine tuning our network with this new augmented dataset leads to noticeable improvements in depth prediction. 
Figure \ref{fig:visualization_aug} illustrates how $8 \times$ data for the fine tuning improves the reconstructions. Notice in particular the improved localization of object edges.
This is particularly encouraging for our stereopsis loss based unsupervised training procedure as its fine tuning only requires a cheap stereo-rig to collect new data in the wild. 
%
For example, we can resort to much larger road scene understanding dataset like cityscapes \cite{Cordts2016Cityscapes} (captured without laser sensor) or 
 a vast collection of 3D movies much like recently published work Deep3D \cite{xie2016deep3d} to repeat this fine-tuning experiment 
for single view depth prediction in the wild.

\begin{table}[!t]
\centering
\caption{ \label{table_soa_comparison}Comparison with state-of-the-art methods on KITTI dataset. }
\begin{scriptsize}
 \begin{tabular}{|c |c|c|c|c|c|c|c|c|c|c|c|c|c|}
 \hline
		    & &  &$log$  & Absolute &Square  &&Accuracies& \\ 
 Methods &Resolution&RMS&RMS& relative& relative&$\delta < 1.25$ &$\delta < 1.25^2$ &$ \delta <1.25^3$\\ \hline

 Ours  L12	&$176\times 608$	&{ 5.285} &{0.282}  &{0.177} &{ 1.169}		&{0.727} &{0.896} &{0.958}
\\ \hline 
 Ours  L12, Aug 8x	&	&\textbf{ 5.104} &{0.273}  &\textbf{0.169} &\textbf{ 1.080}		&\textbf{0.740} &\textbf{0.904} &{0.962}
 \\ \hline \hline
  Mean  		&- &9.635 &0.444  & 0.412 & 5.712 		&0.556 &0.752 &0.870
 \\ \hline
 Make3D \cite{make3dPAMI2009}  		& Dense &8.734 &0.361  & 0.280 & 3.012 		&0.601 &0.820 &0.926 
 \\ \hline
 
 Eigen {\em et al} (c\footnotemark)  \cite{DeptheigenNIPS2014} 	&$28\times 144$ &7.216 &0.273  & 0.194 & 1.531 		&0.679 &0.897 &\textbf{0.967}
 \\ \hline
 Eigen {\em et al} (f)  \cite{DeptheigenNIPS2014}	& $27\times 142$&7.156 &\textbf{0.270}  & {0.190} & 1.515 	&0.692 &{0.899} &\textbf{0.967}
 \\ \hline
 Fayao {\em et al} (pt)  \cite{DepthLiuPAMI2016}	& superpix &7.421 &-  & - & - 		&0.613 &0.858 &0.949
 \\ \hline 
Fayao {\em et al} (ft) \cite{DepthLiuPAMI2016}	& superpix  &7.046 &-  & - & - 		&0.656 &0.881 &0.958
 \\ \hline 
 \end{tabular}
 \end{scriptsize}
\end{table}
\footnotetext{$c$ and $f$ indicates the the coarse and fine networks of \cite{DeptheigenNIPS2014}. Also pt and ft indicates the the pre-trained and fine-tuned networks of \cite{DepthLiuPAMI2016}.}

\begin{figure}[!t]
\centering\includegraphics[width=1\textwidth]{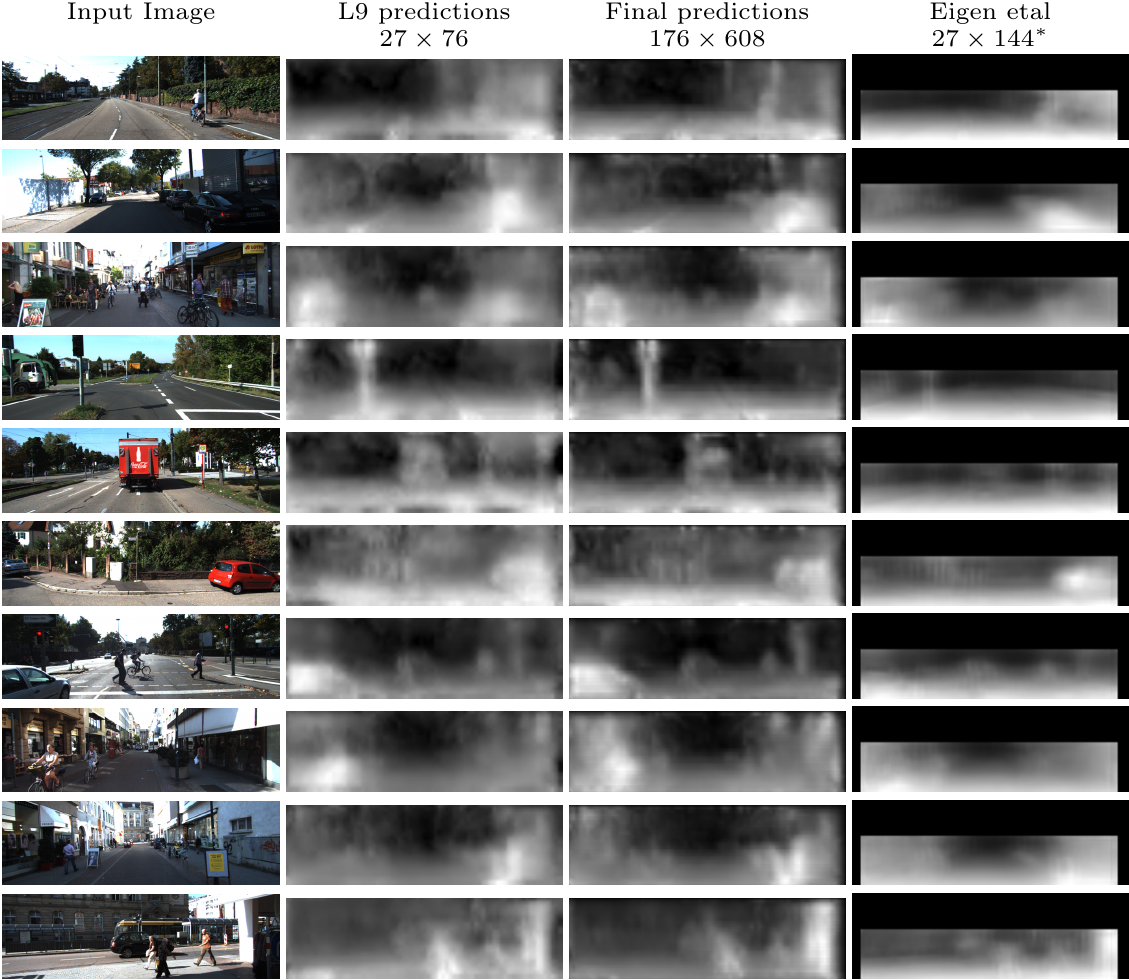} 
\caption{Inverse Depths visualizations. Brighter color means closer pixel.\label{fig:visualization}}
\end{figure}

\subsection{Comparison with State-of-the-art Methods on KITTI Dataset}
In Table \ref{table_soa_comparison}, we compare the performance our network with state-of-the-art single view depth prediction methods \cite{make3dPAMI2009,DeptheigenNIPS2014,DepthLiuPAMI2016}.
Errors for other methods are taken from  \cite{DeptheigenNIPS2014,DepthLiuPAMI2016}. 
%
%
Our method achieves the lowest RMS and Square relative error on the dataset and significantly outperforms other methods for these measures.  
It performs on par with the state-of-the-art methods on all other evaluation measures.
Eigen {\em et al} \cite{DeptheigenNIPS2014} obtains slightly lower error in terms of $log$ RMS compared to ours. However,
as \cite{DeptheigenNIPS2014,DepthLiuPAMI2016} are trained by minimizing  $log$ RMS error with respect to the true depths, we expect the best performance of these methods under same metric.

The most noteworthy point is that our is a completely unsupervised network trained with randomly initialized weights, whereas \cite{DeptheigenNIPS2014} and \cite{DepthLiuPAMI2016}
initialize the networks using Alexnet and VGG-16 respectively, and are supervised.

Figure \ref{fig:visualization} compares the output inverse depthmaps (scaled to [0 1]) for the L9 ($2^{nd}$ column) and L12 ($3^{rd}$ column) layers of the proposed method and \cite{DeptheigenNIPS2014}. 
We appropriately pad the predictions provided by the authors of \cite{DeptheigenNIPS2014} to generate the visualizations at the correct scale. 
It is evident from the figure that both L9 and L12 are able to capture objects that are closer to the camera with significantly more details.
For example, notice the traffic light in Row 4, truck in Row 5 and pedestrians in Row 6 and Row 10; these important scene elements are ``washed out'' in the predictions generated by \cite{DeptheigenNIPS2014}.
Edges are localized more accurately in  L12 results compared to L9. This depicts that even with the simple linear interpolation of the coarse depth estimation, 
the finer image alignment errors are correctly back-propagated leading to the performance boost. Blurred object boundaries in the finer reconstructions
point to well-known limitations of upsampling based approaches which to a certain extent can be addressed with the ‘atrous’ algorithm \cite{chen14semantic}, 
a fully connected CRF \cite{crfasrnn_ICCV2015,koltun2011efficient} or polynomial interpolations replacing simple linear interpolation layers.

In summary, our simple, skinnier network than \cite{DeptheigenNIPS2014} gives on par results without any supervision, and which look visually more appealing.  
Our results could be further refined using better loss functions and replacing linear interpolation filter with a learned CRF. As our method is completely unsupervised, 
it can be trained on theoretically limitless data with deeper networks to capture variation and give depthmaps at full image resolutions.

\subsection{Comparisons with Baseline Supervised Networks and Stereo}
As discussed in Section \ref{sec:intro}, an alternative to our proposal of directly minimizing the 
loss \eqref{eq:LossHS}, would be to train with a standard ``depth loss'' using the output of an off-the-shelf stereo algorithm to generate proxy ground-truth depth for training.
In this section, we substantiate that the proposed autoencoder framework is superior to this alternative approach (which we denote as Stereo $\rightarrow$ CNN).
%
For this purpose, we train the network described in Figure \ref{fig:net_arch}, end-to-end, with least square loss on the disparity difference between CNN prediction and stereo prediction.
\footnote{Much like the log depth, inverse depth parametrization is less prone to the higher depth errors at very distant points and is used successfully
in many stereo \cite{GeigerIJRR2013} and SLAM frameworks \cite{newcombe2011dtam}. 
} 
%

\begin{table}[!t]
\caption{Comparison of proposed auto-encoder framework with a supervised CNN trained on stereo data, and stereo baselines on the KITTI dataset. } \label{tab:stereo_comparison}
\centering
\begin{scriptsize}
 \begin{tabular}{|c |c|c|c|c|c|c|c|c|c|c|c|c|c|}
 \hline
		    & &  &$log$  & Absolute &Square  &&Accuracies& \\ 
 Methods &Coverage&RMS&RMS& relative& relative&$\delta < 1.25$ &$\delta < 1.25^2$ &$ \delta <1.25^3$\\ \hline

 Ours  L12	&{100\%}	&\textbf{ 5.285} &\textbf{0.282}  &\textbf{0.177} &\textbf{ 1.169}		&\textbf{0.727} &\textbf{0.896} &\textbf{0.958}


  \\ \hline 
  HS$\rightarrow$ CNN,$\gamma = .01$   &{100\%}	&6.691 &0.385  & 0.309 & 2.657 		&0.476 &0.750 &0.891
 \\ \hline
   HS$\rightarrow$ CNN  		&{100\%}	&6.292 &0.338  & 0.238 & 1.639 		&0.573 &0.841 &0.941
 \\ \hline

  SGM$\rightarrow$ CNN 			&{100\%}	&5.680 &0.300  & 0.185 & 1.370 		&0.703 &0.886 &0.955 

 \\ \hline \hline \hline
  HS-Stereo, $\gamma = .01$		&\textbf{100\%}	&6.077 &0.381  & 0.299 & 3.264 		&0.677 &0.822 &0.90
  \\ \hline 
   HS-Stereo 				&\textbf{100\%}	&6.760 &0.366  & 0.254 & 4.040 		&0.754 &0.872 &0.928
  \\ \hline 
  SGM-Stereo &{87\% }			&3.030 &0.150  & 0.064 & 0.506 		&0.955 &0.979 &0.989
  \\ \hline
 \end{tabular}
 \end{scriptsize}
\end{table}

\begin{figure}[!t]
 \begin{tabular}{ccc}
 \scriptsize{Input Image} &\scriptsize{SGM-Stereo} & \scriptsize{HS-Stereo}\\
  \includegraphics[width=0.33\textwidth]{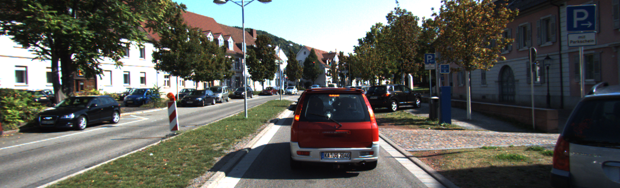}
  &\includegraphics[width=0.33\textwidth]{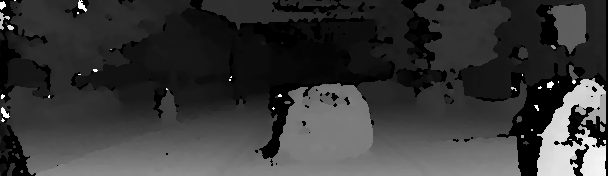}
  &\includegraphics[width=0.33\textwidth]{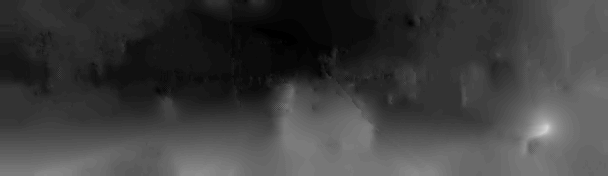}\\

  \scriptsize{Our Predictions} &\scriptsize{SGM$\rightarrow$CNN} & \scriptsize{HS$\rightarrow$CNN}\\
  \includegraphics[width=0.33\textwidth]{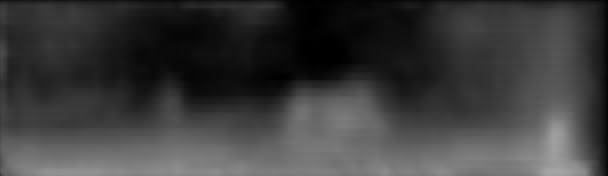}
  &\includegraphics[width=0.33\textwidth]{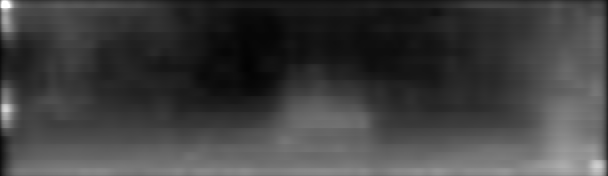}
  &\includegraphics[width=0.33\textwidth]{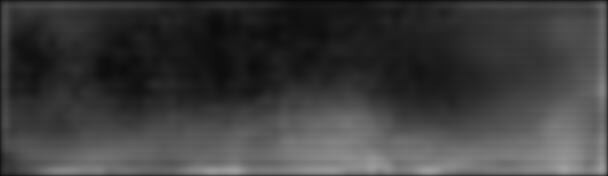}\\
    \scriptsize{Our Errors}  &\scriptsize{SGM$\rightarrow$CNN Errors}& \scriptsize{HS$\rightarrow$CNN Errors}\\
    \includegraphics[width=0.33\textwidth]{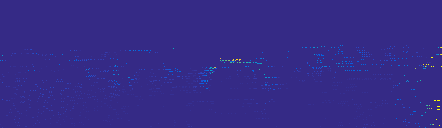}
  &\includegraphics[width=0.33\textwidth]{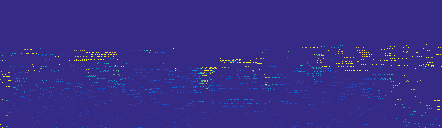}
  &\includegraphics[width=0.33\textwidth]{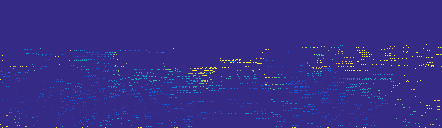}
 \end{tabular}
\caption{Comparing depth predictions baseline stereo methods (top row), with the proposed unsupervised CNN (left column-middle row) and Stereo$\rightarrow$CNN approaches (center/right column-middle row).
Bottom row shows the depth estimation errors as heat-maps for the corresponding methods in middle row.
\label{fig:visualization_Stereo}}
\end{figure}

To generate the stereo prediction, we use a variational Horn-Schunck algorithm.  While this is clearly not a state-of-the-art stereo algorithm, it is a fair baseline since this is the same loss on which we train our photometric loss network. We use the OpenCV implementation, with 6 coarse-to-fine pyramid levels
with scale factor 0.5. To make sure the algorithm converges properly, we increased the number of warp iterations to 1000.
We additionally tried HS $\rightarrow$ CNN with the disparity regularization strength $\gamma = 0.01$ as well, but the results were less accurate.

As shown in Table \ref{tab:stereo_comparison}, depth prediction accuracy of this HS $\rightarrow$ CNN baseline falls significantly short of the proposed framework on all accuracy measures. 
We also incorporate the test-set depth estimation accuracies for the baseline HS-stereo method (which uses both left and right image) for the reference.
A very surprising observation is that our single view depth prediction network works on par with even HS-stereo 
thanks to the common structure present in the road scenes that our network successfully learns. 
Having access to two images, HS-stereo was able to estimate disparity of the closer points
with much more precision but over-reliance on the depth regularization and unawareness of the scene context
results in wrong depths near edges -- where the single view depth estimation even outperforms the HS Stereo. 

In addition to HS-Stereo, we also used Semi Global Matching (SGM) algorithm \cite{hirschmuller2005accurate} to supervise the CNN.
Semi Global matching is known to produce more accurate depths and is an integral part of many of the state-of-the-art stereo algorithms on KITTI stereo dataset \cite{GeigerIJRR2013}.
This stereo method gave very accurate results on the test-set for 87\% of the pixels but left holes in the reconstructions.  
We train SGM $\rightarrow$ CNN by minimizing the sum of least square error for predicted disparities  on the training data, ignoring the points where SGM gave no disparity.
We observed SGM $\rightarrow$ CNN performed on par with the state-of-the-art fully supervised single view depth estimation algorithm but the results were not as accurate as the proposed approach.
We believe that the reason for this was the systematic holes which were left in the SGM-Stereo reconstructions. 

To validate this, in Figure \ref{fig:visualization_Stereo} we analyze if regions with lower depth accuracy of SGM$\rightarrow$CNN coincide with the holes left by SGM-Stereo.
The correlation in errors SGM-Stereo depthmap with that of SGM $\rightarrow$CNN suggests that the supervised training with proxy ground-truth indeed is prone to learn systematic 
errors in the proxy ground truth and advocates need for a more principled integration of a state-of-the-art stereo method with deep learning.
The proposed autoencoder setup 
 is the reasonable first step towards this goal.


\section{Related work}
In this work we have proposed a geometry-inspired unsupervised setup for visual learning, in particular addressing the problem of single view depth estimation.
Our main objective was to address the downsides of training deep networks with large amount of labeled data.
Another body of work which attempts to address this issue is  the set of methods like \cite{HandaPBSC15a,NIPS2014_5548,G_2016_CVPR} which  rely mainly on generating synthetic/semi-synthetic training data 
with the aim to mimic the real world and use it to train deep network in a \textit{supervised} fashion. For example, 
in \cite{NIPS2014_5548}, CNN is used to discriminate a set of surrogate classes where the data for each class is generated automatically from unlabeled images. The network thus learned is shown to perform well 
on the task image classification.  Handa {\em et al} \cite{HandaPBSC15a} learn a network for semantic segmentation using synthetic data of indoor scenes and show that the network can 
generalize well on the real-world scenes. 
Similarly, \cite{G_2016_CVPR} employs a CNN to learn local image descriptors where the correspondences between the patches are obtained using a multi-view stereo algorithm.


Recently, many methods have used CNN to
learn good visual features for matching patches which are sampled from stereo datasets like KITTI \cite{stereomatchingCVPR2015,stereocostICCV2015},
and match these features while doing classical stereo to achieve state-of-the-art depth estimation. 
These methods are reliant on local matching and lose global information about the scene; furthermore they use ground-truth.
But their success is already an indicator that a joint visual learning and depth estimation approach like ours could be extended at the test time to use a pair of images. 

There have been few works recently that approach the problem of novel view synthesis with CNN \cite{xie2016deep3d,flynn2015deepstereo}.
Deep stereo \cite{flynn2015deepstereo} uses a large set of posed images to learn a CNN that can interpolate between the set of input views that are separated by a wide baseline. 
A concurrent work with ours, \cite{xie2016deep3d} addresses the problem of generating 3D stereo pairs from 2D images. 
It employs a CNN to infer a soft disparity map from a single view image which in turn is used to render the second view. 
Although, these methods generate depth-like maps as an intermediate step in the pipeline, their goal however is to generate new views and hence do not evaluate the computed depth maps

Using camera motion as the information for visual learning is also explored in the works like \cite{learningtoseeICCV2015,LongKAL16} which directly 
regress over the 6DOF camera poses to learn a deep network which performs well on various visual tasks. In contrast to that work, we train our CNN for a more generic task of synthesizing image
and get the state-of-the-art single view depth estimation. 
It will be of immense interest to evaluate the quality of the features learned with our framework on other semantic scene understand tasks.

\section{Conclusions}

In spite of the enormous growth and success of deep neural networks for a variety of visual tasks, an abiding weakness is the need for vast amounts of annotated training data.  We are motivated by the desire to build systems that can be trained relatively cheaply without the need for costly manual labeling or even trained on the fly.  To this end we have presented the first convolutional neural network for single-view depth estimation  that can be trained end-to-end from scratch, in a fully unsupervised fashion, simply using data captured using a stereo rig.  We have shown that our network trained on less than half of the KITTI dataset gives comparable performance to the current state-of-the-art supervised methods for single view depth estimation.

Various natural extensions to our work present themselves.  
Instead of training on KITTI data (which is nevertheless convenient because it provides a clear baseline)
we aim to train on a continuous feed from a stereo rig ``in the wild'', and to explore the effect on accuracy
by augmenting the KITTI data with new stereo pairs. Furthermore, as intimated in the Introduction, our method
is not restricted to stereo pairs, and a natural extension is to use a monocular SLAM system to compute camera motion,
and use this known motion within our autoencoder framework; here the warp function is slightly more complex than for rectified stereo,
but still well understood.  The resulting single-view depth estimation system could be used for bootstrapping structure,
or generating useful priors on the scene structure that capture much richer information than typical continuity or smoothness assumptions.
It also seems likely that the low-level features learned by our system will prove effective for other tasks such as classification,
in a manner analogous to \cite{learningtoseeICCV2015,NIPS2014_5548}, but this hypothesis remains to be proven experimentally.

\noindent\textbf{Acknowledgement}: This research was supported by the Australian Research Council through the Centre of Excellence in Robotic Vision, CE140100016, and through Laureate Fellowship FL130100102 to IDR.

\clearpage

\bibliographystyle{splncs03}
\bibliography{egbib}
\end{document}